\documentclass[letterpaper, conference]{IEEEtran}
\IEEEoverridecommandlockouts
\usepackage{cite}
\usepackage{svg}
\usepackage{amsmath,amssymb,amsfonts}
\usepackage{graphicx}
\usepackage{textcomp}
\usepackage{xcolor}
\usepackage{algorithm, algorithmic}
\usepackage{subcaption}
\usepackage{hyperref}

\usepackage{array}
\newcommand{\PreserveBackslash}[1]{\let\temp=\\#1\let\\=\temp}
\newcolumntype{C}[1]{>{\PreserveBackslash\centering}p{#1}}

\def\BibTeX{{\rm B\kern-.05em{\sc i\kern-.025em b}\kern-.08em
    T\kern-.1667em\lower.7ex\hbox{E}\kern-.125emX}}
\begin{document}

\title{ DSOR: A Scalable Statistical Filter for Removing Falling Snow from LiDAR Point Clouds in Severe Winter Weather }

\author{\IEEEauthorblockN{Akhil Kurup}
\IEEEauthorblockA{\textit{Electrical and Computer Engineering} \\
\textit{Michigan Technological University}\\
Houghton, MI, USA \\
orcid: 0000-0003-2262-9777}
\and
\IEEEauthorblockN{Jeremy Bos}
\IEEEauthorblockA{\textit{Electrical and Computer Engineering} \\
\textit{Michigan Technological University}\\
Houghton, MI, USA \\
orcid: 0000-0002-9625-2274}
}

\maketitle

\begin{abstract}
For autonomous vehicles to viably replace human drivers they must contend with inclement weather. Falling rain and snow introduce noise in LiDAR returns resulting in both false positive and false negative object detections. In this article we introduce the Winter Adverse Driving dataSet (WADS) collected in the snow belt region of Michigan's Upper Peninsula. WADS is the first multi-modal dataset featuring dense point-wise labeled sequential LiDAR scans collected in severe winter weather; weather that would cause an experienced driver to alter their driving behavior. We have labelled and will make available over 7 GB or 3.6 billion labelled LiDAR points out of over 26 TB of total LiDAR and camera data collected. We also present the Dynamic Statistical Outlier Removal (DSOR) filter, a statistical PCL-based filter capable or removing snow with a higher recall than the state of the art snow de-noising filter while being 28\% faster. Further, the DSOR filter is shown to have a lower time complexity compared to the state of the art resulting in an improved scalability.

Our labeled dataset and DSOR filter will be made available at \href{https://bitbucket.org/autonomymtu/dsor_filter}{https://bitbucket.org/autonomymtu/dsor\_filter}.

\end{abstract}

\begin{IEEEkeywords}
Autonomous Driving, Adverse Weather Dataset, Point Cloud De-noising Filter
\end{IEEEkeywords}

\section{Introduction}
\label{sec:introduction}
\begin{figure}[htbp]
\centering
\includegraphics[width=0.434\textwidth]{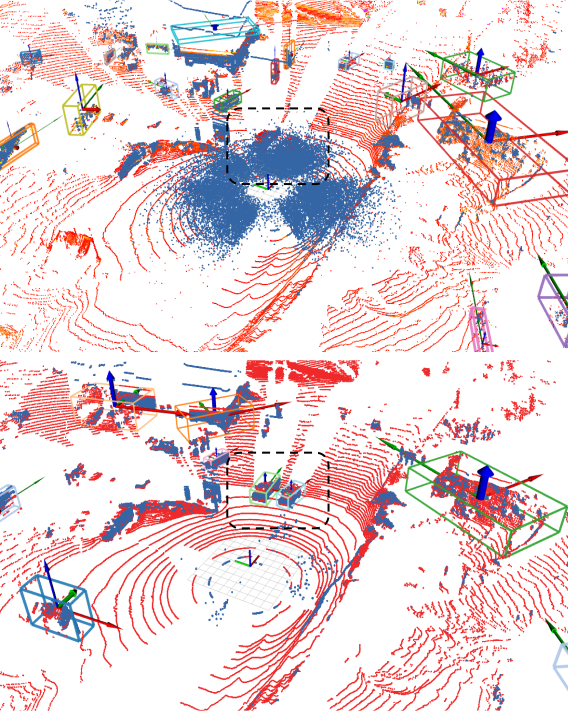}
\caption{Part of our Winter Adverse Driving dataSet (WADS) showing moderate snowfall 0.6 in/hr (1.5 cm/hr). (top) Clutter from snow particles obscures LiDAR point clouds and reduces visibility. Here, two oncoming vehicles are concealed by the snow. (bottom) Our DSOR filter is faster at de-noising snow clutter than the state of the art and enables object detection in moderate and severe snow.}
\label{fig:clutter_clearing}
\end{figure}

Autonomous Vehicle (AV) perception systems often rely on Light Detection And Ranging (LiDAR) sensors for perception, mapping, and localization. Precipitation such as rain and snow can reduce the performance of these systems\cite{park2020}. LiDAR sensors are particularly affected due to their inherent beam divergence and short pulse duration. For many AV focused LiDAR sensors, snowflake detection noise is concentrated near the sensor and manifests as a clutter-noise as seen in Fig. \ref{fig:clutter_clearing}. Snow clutter noise can introduce false detections as well obscure important obstacles leading to critical false-negative detections \cite{bos2020autonomy}. As an example, Fig. \ref{fig:clutter_clearing} (top) shows a LiDAR point cloud captured in Hancock, MI on the 12\textsuperscript{th} of February 2020 with a snow rate of 0.6 in/hr (1.5 cm/hr). Two oncoming vehicles are obscured by the snow and evade Autoware's Euclidean clustering for object detection \cite{autoware}.

\begin{table*}[t]
  \centering
  \caption{Publicly available datasets with annotated LiDAR scans. WADS is the first dataset to feature dense point-wise labeled LiDAR scans in severe winter weather.}
  \label{tab:datastes}
  \renewcommand{\arraystretch}{1.1}
  \begin{tabular}{ l*{6}{c} }
  \hline
  Dataset & LiDAR & Labeling & Inclement Weather\\
  \hline
  \hline
  SemanticKITTI\cite{behley2019semantickitti}	& 64 channel & point-wise & No\\
  nuScenes-lidarseg\cite{caesar2020nuscenes}	& 32 channel & point-wise & No\\
  ApolloScape \cite{huang2020} & 64 channel & point-wise & No\\
  DENSE\cite{bijelic2020}	& 32 \& 64 channel & bounding boxes & Yes\\
  CADC\cite{pitropov2020} & 32 channel & bounding boxes & Yes\\
  \hline
  \textbf{WADS} & \textbf{64 channel} & \textbf{point-wise} & \textbf{Yes}\\
  \hline
  \end{tabular}
\end{table*}

Precipitation is often hard to predict and severe events are infrequent. Populated North American cities such as Chicago, Detroit, Minneapolis, etc. can receive over 1-inch (2.5 cm) of snow per hour during severe winter weather events. To enable universal adaptation of AV's, perception systems must be able to reliably operate in conditions like these that may arise unexpectedly or would be tolerated by an experienced human driver. Neural Networks (NN) trained on large datasets, fail to perform well in winter driving conditions; lack of available winter weather datasets has been cited as a likely reason \cite{pfeuffer2018optimal}. Michigan's Keweenaw peninsula frequently receives on average over 200 in. (500 cm) of snow annually. Here, blowing snow can result in both intermittent and persistent white out conditions where visibility is near zero. The frequency of such adverse weather in this area allows reliable collection of large winter driving data featuring extreme snow events \cite{bos2020autonomy}. In this paper, we introduce the first point-wise labeled dataset featuring moderate to severe winter driving conditions collected in sub-urban environments. In the context of this work we refer to severe winter weather as conditions that would results in a change in behavior of an experience human driving due to loss of visibility or uncertainty in road surface conditions. Our Winter Adverse Driving dataSet (WADS), aptly named after the largest dorm on Michigan Tech's campus, features over 7GB of labeled LiDAR point clouds featuring active falling snow as well as accumulated snow on the side of the roads from vehicle movement and snow removal. Point-wise labels enable every point in a scan to have a unique class facilitating capture of fine details such as individual snowflakes. Such a labeled dataset dedicated to winter driving will propel development of AV's in adverse weather and pave the way for their universal adaptation.

A  common technique for dealing with noise in LiDAR point clouds is a filtering step prior to a detection stage. Generalized filters, however, do not account for the distribution of the snow and risk losing important keypoints in the environment. LiDAR returns are volumetrically non-uniform resulting in a point-density that decreases with increasing range. Noise introduced by falling snow, shows the same non-uniformity and follows a log-normal distribution \cite{michaud2015}. To address this limitation, we present the Dynamic Statistical Outlier Removal (DSOR) filter to remove snow from LiDAR point clouds. Our solution dynamically changes the filter threshold with range to remove most of the snow and outlier noise while preserving environmental features. Fig. \ref{fig:clutter_clearing} (bottom) shows most of the snow has been cleared by our DSOR filter. We compare our filter to the Dynamic Radius Outlier Removal (DROR) filter \cite{charron2018denoising}, which we consider to be the state of the art filter for snow de-noising. The proposed DSOR filter is shown to perform 28\% faster than the DROR filter with a 4\% increase in the filter recall.

The contributions of this paper can be summarized as follows:
\begin{itemize}
    \item We introduce the first dense point-wise labeled Winter Adverse Driving dataSet (WADS) featuring moderate to severe snow.
    \item We introduce the DSOR filter to de-noise point clouds corrupted by falling snow.
    \item We show that the DSOR filter outperforms the state of the art in recall and has a lower time complexity resulting in improved scalability. %feel free to remove this%
\end{itemize}

\section{Related Work}
\label{sec:relatedwork}
Automotive LiDARs can be adversely affected by weather related scattering and absorption effects. Rain adds noise to point clouds severely degrading performance of object detection algorithms \cite{xu2021spg}. Falling snow in winter can be detected as diffuse, solid objects, often smaller than the laser beam cross section. Rasmussen \textit{et al.} \cite{rasmussen1999estimation} show that small snow particles act as Lambertian targets and create significant backscattering. To better understand the interaction of snow with automotive 3D LiDAR, Roy \textit{et al.} \cite{roy20} modeled the interaction between snowflakes and laser pulses. They show that snow detections are concentrated near the sensor and detection statistics are governed by the characteristics of the laser beam and rate of precipitation. Michaud \textit{et al.} \cite{michaud2015} show that the probability of detecting snow in automotive 3D LiDAR follows a log-normal distribution with distance from the sensor. %Not sure if this is true but it was missing some quantity 

\subsection{Datasets}
\label{subsec:litt_datasets}

Over the past decade, several annotated datasets have been released with LiDAR scans to aid with the development of AV perception tasks \cite{xie2020}. Few include multi sensor winter weather data with extreme events necessary for training NN models. The lack of inclement weather data has been addressed in some literature by adding artificial corruption to existing datasets \cite{sakaridis2018, heinzler_cnn-based_2020}. Pfeuffer and Dietmayer \cite{pfeuffer2018optimal} show that models trained on large datasets such as KITTI \cite{geiger2013vision}, fail to perform well when artificial rain or snow is introduced implying that availability of adverse data takes precedence over the size of the dataset. 

The KITTI\cite{geiger2013vision} and nuScenes\cite{caesar2020nuscenes} datasets provide LiDAR scans annotated with bounding boxes while SemanticKITTI \cite{behley2019semantickitti} and nuScenes-lidarseg datasets provide point-wise annotations. These datasets, however, provide no data in inclement weather. The ApolloScape dataset \cite{huang2020} includes LiDAR scans with a semantic mask to extract point-wise annotations. The DENSE dataset \cite{bijelic2020} includes rain, fog, and snow but the conditions do not rise to the level of being severe. Moreover, annotations in DENSE are limited to bounding boxes and not point-wise labels. The CADC dataset \cite{pitropov2020} includes adverse weather data collected in Canada with bounding boxes around vehicles and pedestrians. Our dataset provides point-wise annotations and have been collected in consistent severe conditions. Table \ref{tab:datastes} provides an overview of relevant datasets compared with ours.

\subsection{Filters}
\label{subsec:litt_filters}
Noise introduced by snow particles in LiDAR scans can be filtered out in a number of ways; \cite{park2020} and \cite{roy20} introduce intensity based filters. These filters assume LiDAR returns from snowflakes will have a lower intensity and remove snow noise by intensity-value thresholding. However, intensity is dependant both on the laser wavelength and target reflectance and, as shown in \cite{rasmussen1999estimation}, backscatter from snow particles near the sensor can be significant. To overcome this limitation, Park \textit{et al.} \cite{park2020} propose a second clustering stage based on point density to preserve objects of interest. However, in heavy snow with a high volume density clustering techniques will classify snow as objects of interest.

Another filtering technique is to apply image processing on LiDAR point clouds. Here, point clouds are converted to a 2D representation and techniques like Gaussian filters or their derivatives are applied to remove noise. However, as shown by Charron \textit{et al.} \cite{charron2018denoising}, 3D point clouds are sparse and such approaches not only fail to effectively remove snow but also have the negative effect of smoothing edges in keypoint features. Duan \textit{et al.} \cite{duan2020lowcomplexity} use Principal Component Analysis (PCA) to convert 3D point clouds to 2D and apply DBSCAN clustering to find sparse regions and remove them. However, as mentioned previously, clustering techniques fail at high snowfall rates.

% Neural Networks have also been used for filtering. A CNN based de-noising approach has been introduced by Heinzler \textit{et al.} \cite{heinzler_cnn-based_2020}. They emulate rain and fog in a climate chamber and also augment existing datasets with artificial noise. Their model has shown to work well on rain and fog but they have not evaluated in the presence of falling snow.

Filtering can also be applied to the 3D points directly. The Point Cloud Library (PCL) \cite{rusu2011pcl} includes general purpose filters such as the voxel grid filter, Statistical Outlier Removal (SOR) filter and Radius Outlier Removal (ROR) filter; these are, however, not designed for snow. 
% The SOR filter classifies points as outliers if the standard deviation of their mean distance to their neighbors is below a user defined threshold. The ROR filter computes the number of neighbors for every point within a specified radius and removes points less than a user defined minimum. 
They fail to account for the non-uniformity of the point clouds and hence Charron \textit{et al.} \cite{charron2018denoising} propose the Dynamic Radius Outlier Removal (DROR) filter for removing snow corruption. Their filter dynamically changes the search radius with range to effectively remove snow. In this work, we compare our DSOR filter with the DROR filter, which we consider to be the state of the art.

\begin{figure}[htbp]
\centering
\includegraphics[width=0.475\textwidth, height=0.25\textwidth]{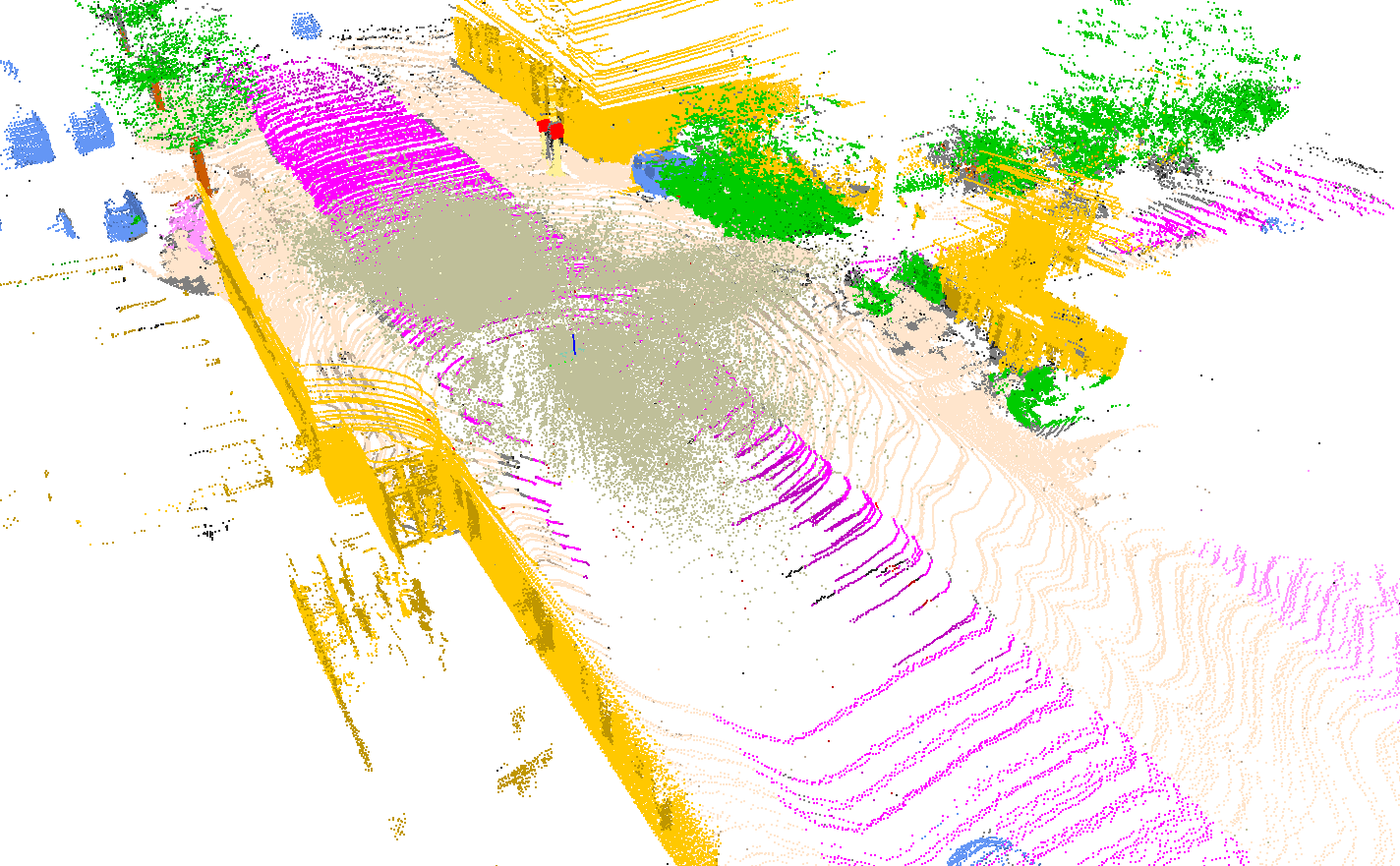}
\caption{A labeled sequence from the WADS dataset. Every point has a unique label and represents one of 22 classes. Active falling snow (beige) and accumulated snow (off-white) are unique to our dataset.}
\label{fig:dataset_sample}
\end{figure}

\section{The WADS Dataset}
\label{sec:dataset}
This work introduces point labels for our winter LiDAR dataset \cite{bos2020autonomy}. Captured data has been split into sequences of approximately 100 scans each. Each set has sequential scans to further development of algorithms using spatial information. Multiple suburban locations have been captured, including 2-lane highways, residential areas and parked as well as moving vehicles. A total of 26 TB of multi-modal data have been captured. We apply point-wise labels to the LiDAR scans and will make available over 7 GB of data amounting to over 3.6 billion labeled points. Simultaneous Long-Wave Infra-Red (LWIR) and visible color camera imagery are also available \cite{bos2020autonomy} but not considered here.

\begin{figure*}[htbp]
\centering
\includegraphics[width=0.75\textwidth]{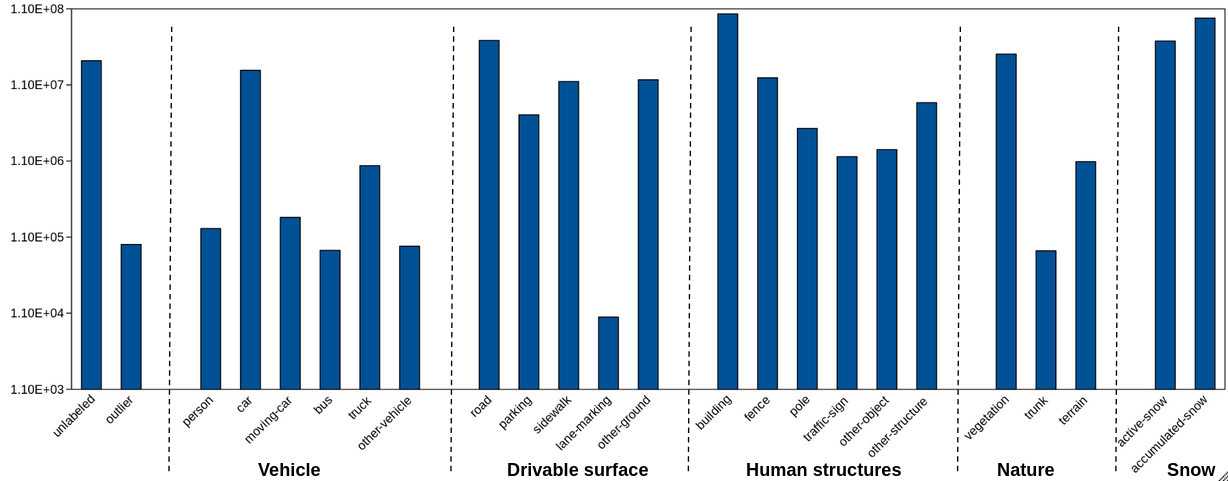}
\caption{Distribution of classes in the WADS dataset. Scenes from sub-urban driving including vehicles, roads, and man-made structures are included. Two novel classes: falling-snow and accumulated-snow are introduced to improve AV perception in adverse winter weather.}
\label{fig:distribution_ratios}
\end{figure*}

\subsection{Labeling}
\label{subsec:labeling}
Bounding boxes provide vector annotations and often include undesired background objects which can be detrimental for AV perception tasks such as semantic segmentation. We have opted for point-wise labels as they are more precise and enable fine details in the environment to be highlighted, such as individual snowflakes. To maintain compatibility with existing systems and ensure adoption of inclement weather data into existing frameworks, we adopt the popular KITTI format \cite{Geiger2013IJRR}. We leverage the point-cloud labeling tool introduced by Behley \textit{et al.} \cite{behley2019semantickitti}. This is a manual labeling tool that can load point clouds either as multiple scans for quick labeling or as single scans for fine details. Fig. \ref{fig:dataset_sample} shows a labeled scan from our WADS dataset.

\subsection{Statistics}
\label{subsec:statistics}
Every point in a LiDAR scan has been labeled into one of 22 classes as shown in Fig. \ref{fig:distribution_ratios}. Over 7 GB or 3.6 billion points have been labeled in all. Here, classes are grouped into categories for easy viewing. It can be seen that the majority of labeled points lie in urban driving scenarios with roads, buildings and various types of vehicles representing most of our labeled data.

In addition to these classes, we introduce two new labels to represent snow. \texttt{active-snow} captures falling snow particles and associated clutter noise in a LiDAR return. \texttt{accumulated-snow} captures snow that builds up on the sides of drive-able surfaces from vehicle traffic and snow removal. Accumulated snow often changes, sometimes through a day, enough to confuse feature-based algorithms. Access to such data will be useful for AV tasks like object detection, localization and mapping, and semantic and panoptic segmentation in adverse weather.

\section{Filter Implementation}
\label{sec:filter_implementation}
\begin{figure*}
\centering
\includegraphics[width=\textwidth]{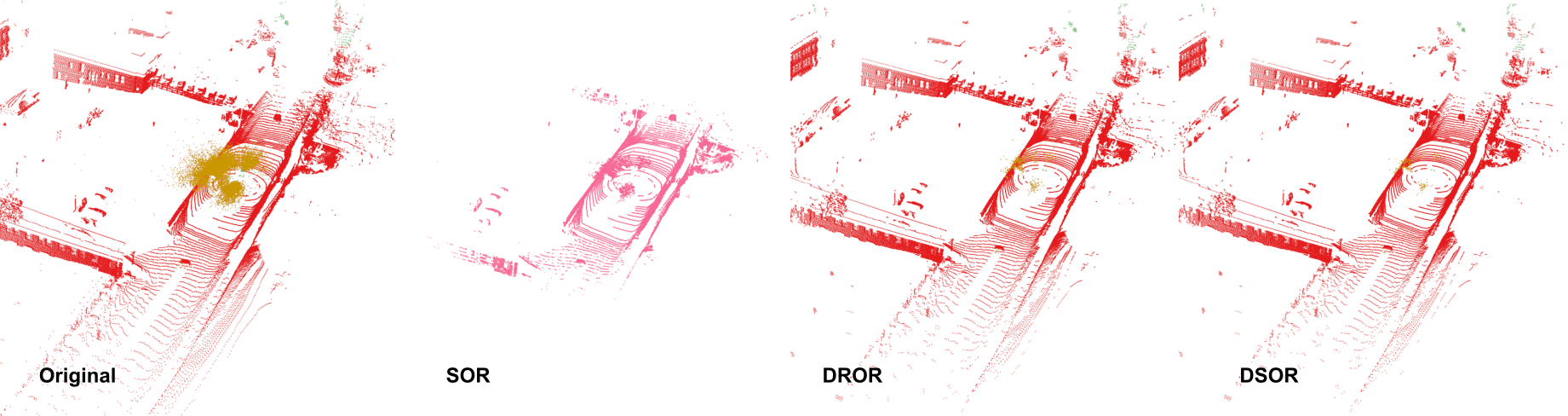}
\caption{Qualitative comparison of the SOR filter, state of the art DROR filter and our DSOR filter (left to right). The original point cloud shows snow clutter (orange points) that degrades LiDAR perception. The DSOR filter removes more snow compared to both the SOR and DROR filters and preserves most of the environmental features.}
\label{fig:qualitative_comparison}
\end{figure*}

PCL's SOR filter is a general noise removal filter widely used for cleaning point clouds. It does not account for the non-uniform distribution of the point cloud and, when applied to a scan with falling snow, fails to remove it as seen in Fig. \ref{fig:qualitative_comparison} (second from left). 

The state of the art DROR filter applies a threshold to the mean distance of points to their neighbors in a given radius to remove sparse snowflakes. To address changing LiDAR point spacing with distance, the DROR filter changes the search radius as the distance increases from the sensor. The DROR filter achieves a high accuracy but fails to produce a clean point cloud at high snowfall rates as shown in Fig. \ref{fig:qualitative_comparison} (second from right).

% \subsection{DROR Filter}
% \label{subsec:dror_filter}
% The DROR filter is an extension to PCL's Radius Outlier Removal (ROR) filter and was developed to address its uniformity assumption \cite{charron2018denoising}. The point cloud is first loaded into a k-d tree and the mean distance from every point to its neighbors within a radius is calculated. The ROR filter applies a simple threshold based on the number of points in this radius. To address point spacing at a distance, the DROR filter changes the search radius (\textit{SR}), given by equation \ref{eqn:dror}, as the distance of a point ($r_p$) increases from the sensor. Here $\beta$ and $\alpha$ are constants to scale the search radius and account for the angular resolution of the LiDAR respectively.
% \begin{equation}
%     SR = \beta \times r_p \times \alpha
%     \label{eqn:dror}
% \end{equation}

% The DROR filter achieves a high accuracy for snow removal and is considered the state of the art filter in this field. It however, does not account for the statistical distribution of snow and fails to produce a clean point cloud in white out conditions as shown in Fig. \ref{fig:qualitative_comparison} (second from right). We show that our proposed DSOR filter produces a cleaner point cloud and also performs faster than the DROR filter.

\begin{algorithm}
    \caption{Dynamic Statistical Outlier Removal (DSOR) filter}
    \label{algo:dsor}
    \begin{algorithmic}[1]
        \renewcommand{\algorithmicrequire}{\textbf{Input:}}
        \REQUIRE Point Cloud (\textbf{P}) = ${p_i}$, $i = 1,...,N$; $p_i=(x_i, y_i, z_i)$\\
        \textbf{k} = minimum number of nearest neighbors\\
      \textbf{s} = multiplication factor for standard deviation\\
      \textbf{r} = multiplication factor for range\\
        \item[]
        \renewcommand{\algorithmicensure}{\textbf{Output:}}
        \ENSURE  Filtered Point Cloud (\textbf{F}) = ${f_i}$, $i = 1,...,N$; $f_i=(x_i, y_i, z_i)$\\
        \item[]
        \STATE $P \gets KdTree$
        \FOR {$p_i \in P $} 
        \STATE $ mean\_distances \gets nearestKSearch(\textbf{k})$
    \ENDFOR
    \item[]
    \STATE calculate: mean $(\mu) \gets mean\_distances$
    \STATE calculate: standard deviation $(\sigma) \gets mean\_distances$
    \STATE calculate: global threshold $(T_g) \gets \mu + (\sigma \times \textbf{s})$
        \item[]
        \FOR {$p_i \in P $}
            \STATE $distance \gets \sqrt{x_i^2 + y_i^2 + z_i^2}$
        \STATE calculate: dynamic threshold $(T_d) \gets T_g \times \textbf{r} \times distance $
            \IF {$ mean\_distances < T_d $}
            \STATE $f_i \gets p_i$ \COMMENT {Inliers}
            \ENDIF
        \ENDFOR
        \item[]
        \RETURN{} \textbf{F}
    \end{algorithmic} 
\end{algorithm}

\subsection{Proposed DSOR Filter}
\label{subsec:dsor_filter}
The DSOR filter is an extension of PCL's SOR filter, designed to address the inherent non-uniformity in point clouds. Algorithm \ref{algo:dsor} helps visualize our approach. The point cloud is first loaded into a k-d tree and the k-nearest neighbor search is applied for every point to accumulate user provided \textit{k} neighbors. The mean ($\mu$) and standard deviation ($\sigma$) of the mean distances are calculated to compute a global threshold given by equation \ref{eqn:tg}.
\begin{equation}
    T_g = \mu + (\sigma \times constant)
    \label{eqn:tg}
\end{equation}

To account for the non-uniform spacing of points, the distance of every point from the sensor is calculated and a new \textit{dynamic threshold} ($T_d$) is computed using equation \ref{eqn:td}
\begin{equation}
    T_d = T_g \times r \times distance
    \label{eqn:td}
\end{equation}

Points with mean distances less than $T_d$ are considered to be outliers and removed. Here, \textit{r} is a constant, a multiplicative factor for point spacing. Increasing \textit{r} will increase $T_d$ and will result in the filter rejecting fewer points, whereas, reducing \textit{r} will decrease $T_d$ resulting in a more aggressive filter. The DSOR filter, when applied to a snow corrupted point cloud, produces the cleanest point cloud as compared to the other approaches as shown in Fig. \ref{fig:qualitative_comparison} (right most).

\section{Evaluation and Results}
\label{sec:results}
We apply the SOR, DROR and DSOR filters on data collected during heavy snow events with severely reduced visibility (severe conditions). All of the filters have been implemented on an Intel® Core™ i7-9750H CPU with 32 GB of RAM. In this section we compare the performance of these filters using both qualitative as well as quantitative results. Qualitative results include both visual performance of the filters and distribution of filtered points as a function of range. To quantify the results, the precision and recall as well as the filtering time have been considered. All results have been averaged over 100 point clouds with falling snow.

\begin{figure}[H]
     \centering
     \begin{subfigure}{0.475\textwidth}
         \centering
         \includegraphics[width=\textwidth]{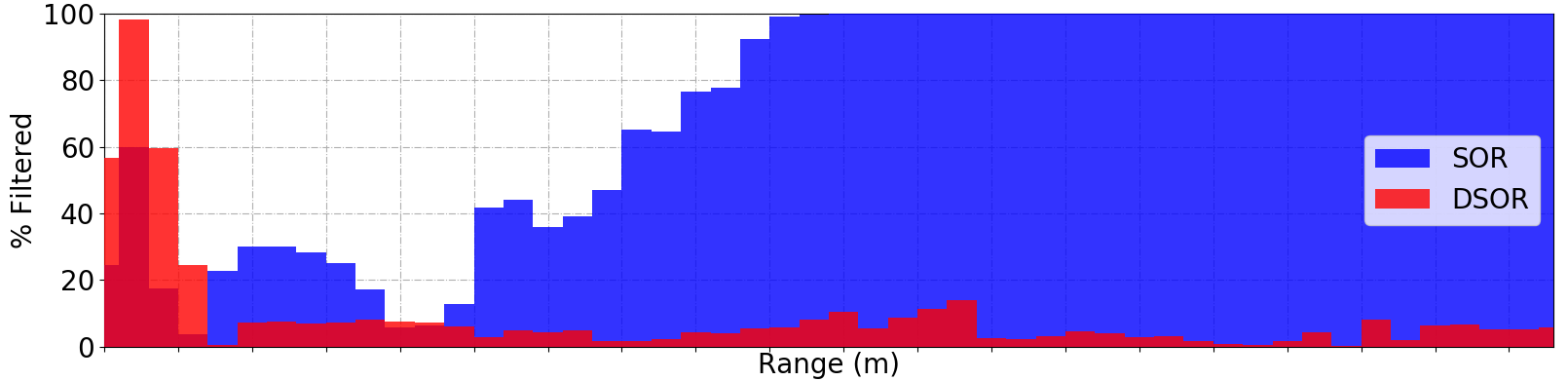}
         \caption{SOR vs. DSOR filter}
         \label{fig:filtered_vs_range_stacked_a}
     \end{subfigure}
     \hfill
     \begin{subfigure}{0.475\textwidth}
         \centering
         \includegraphics[width=\textwidth]{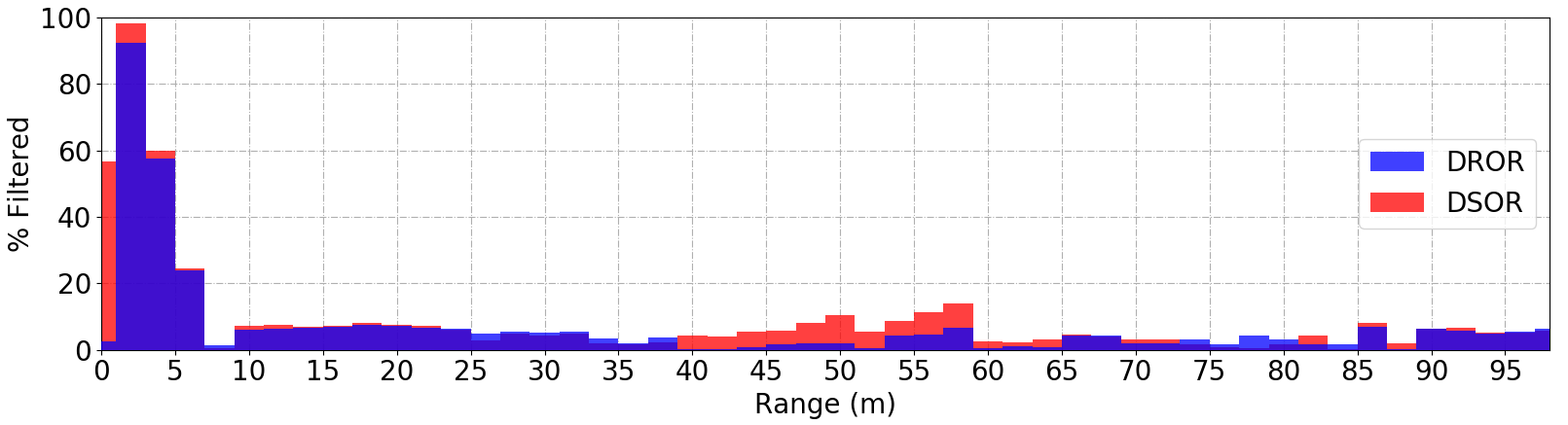}
         \caption{DROR vs DSOR filter}
         \label{fig:filtered_vs_range_stacked_b}
     \end{subfigure}
     \caption{Percentage of filtered points as a function of range (averaged over 100 point clouds). The DSOR filter outperforms both the SOR and DROR filters in ranges $<$ 20m where most of the snow is concentrated.}
     \label{fig:filtered_vs_range_stacked}
\end{figure}

\subsection{Qualitative evaluation}
\label{subsec:qualitative_evaluation}
Fig. \ref{fig:qualitative_comparison} shows a qualitative comparison of the original point cloud, corrupted by snow, passed through different filtering techniques. The SOR filter fails to remove most of the snow and removes key environmental features. The DROR filter does removes most of the snow while preserving environmental features. The DSOR filter removes even more snow detections from the point cloud while preserving key points removed by the SOR filter.

\subsection{Quantitative evaluation}
\label{subsec:quantitative_evaluation}
To understand the visual performance of the filters, the percentage of filtered points as a function of range have been plotted in Fig. \ref{fig:filtered_vs_range_stacked}.

\underline{DSOR vs SOR filter}: 
Fig. \ref{fig:filtered_vs_range_stacked_a} shows that the DSOR filter removes significantly more points than the SOR filter in the first 20 meters where most of the LiDAR interaction with snow particles occurs. Beyond 20 meters, the DSOR filter removes significantly fewer points than the SOR filter thus preserving key environmental features.

\underline{DSOR vs DROR filter}: 
Fig. \ref{fig:filtered_vs_range_stacked_b} shows that in the first 20 meters the DSOR filter also removes more points than the DROR filter. However, beyond 20 meters, the DSOR filter removes more points potentially leading to a lower precision at range.

\begin{figure}[htbp]
\centering
\includegraphics[width=0.475\textwidth]{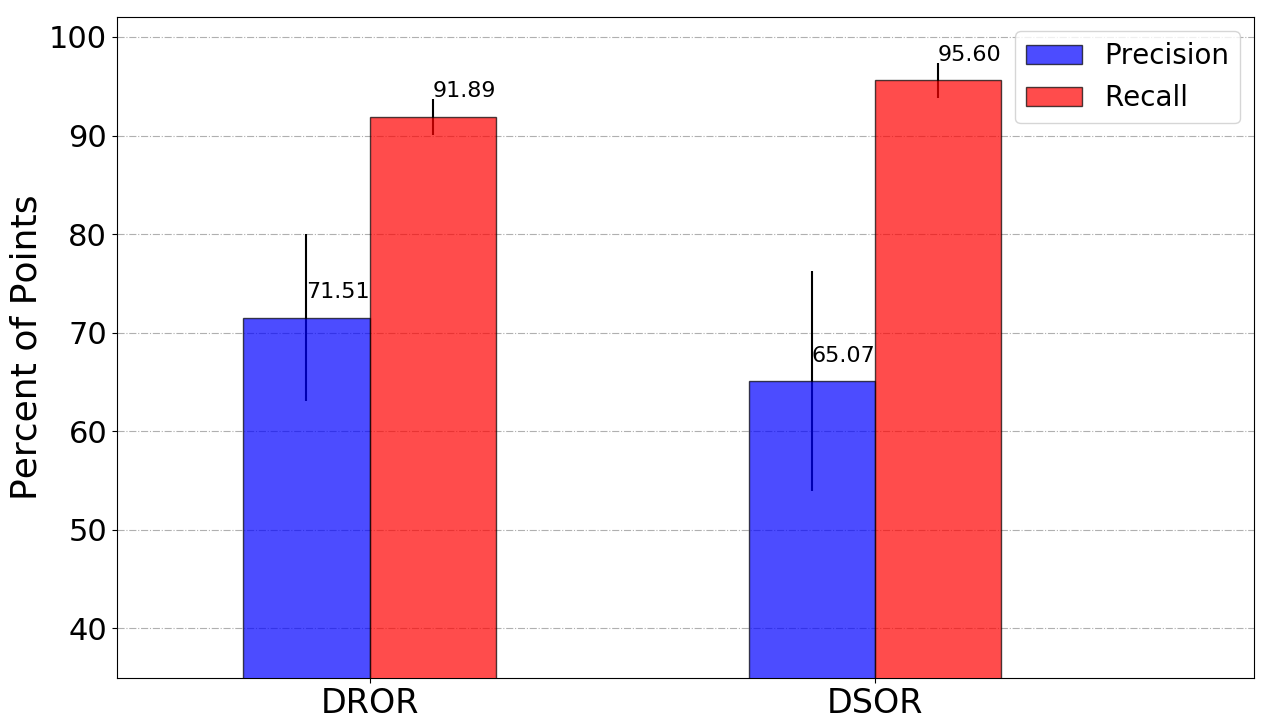}
\caption{The DSOR filter accurately filters out more snow than the DROR filter and achieves a higher recall. Note that the y-axis starts from 40 to better highlight differences between the filters.}
\label{fig:filter_accuracies}
\end{figure}

\subsection{Precision and Recall}
\label{subsec:precision_recall}
In the context of this work, the aim of a filter must be to remove as much snow as possible (recall) and leave other points untouched (precision). It is possible to filter out all of the snow (recall of 100\%) at the expense of environmental features (low precision) using any filter. This, however, is not the aim of this work. Both the DSOR filter and the DROR filter have been tuned to filter most of the snow and preserve as much of the environment as possible as shown in Fig. \ref{fig:qualitative_comparison}.

As shown in Fig. \ref{fig:filter_accuracies}, the DSOR filter achieves a greater recall (95.6\%) than the DROR filter (91.9\%). This shows that the DSOR filter removes more of the snow corruption correctly than the DROR filter. The DROR filter, however, achieves a greater precision (71.5\%) at filtering snow as compared to the DSOR filter (65.1\%). This shows that the DSOR filter removes more non-snow points or environmental features than the DROR filter. From Fig. \ref{fig:filtered_vs_range_stacked_b} it is evident that the loss in precision for the DSOR filter stems from removing points beyond 20 meters. At this range, we observe that features such as tree-tops, poles and overhanging power lines have been filtered out. Such features are sparse, and in the winter can return just a few points. Decrease in precision could also stem from removal of noisy measurements.

\begin{table}[htbp]
  \centering
  \caption{Execution time of the DSOR filter compared with the SOR and DROR filters (averaged over 100 point clouds).}
  \label{tab:filtering_time}
  \renewcommand{\arraystretch}{1.15}
  \begin{tabular}{ |C{1.25cm}|C{1.25cm}|C{1.25cm}|C{1.25cm}| }
  \hline
  & \multicolumn{2}{c|}{Execution Time} & DSOR\\
  Filter &  mean & std. dev. & Speedup\\
  &  (ms) & (ms) & (ms)\\
  \hline
  \hline
  \textbf{DSOR} & \textbf{369.68} & 18.1 & \textbf{-} \\
  \hline
  SOR & 371.98 & 19.2  & \textbf{2.3}\\
  \hline
  DROR & 510.55 & 33.9  & \textbf{140.87}\\
  \hline
  \end{tabular}
\end{table}

% \begin{figure}[htbp]
% \centering
% \includegraphics[width=0.475\textwidth]{figures/filter_times_dsor_vs_sor.png}
% \caption{Our DSOR filter is 2 ms faster than PCL's SOR filter (averaged over 100 point clouds).}
% \label{fig:filter_times_dsor_vs_sor}
% \end{figure}

\subsection{Filter Rate}
\label{subsec:filter_rate}
Table \ref{tab:filtering_time} compares the time taken to filter a point cloud by the DSOR, SOR and DROR filters. Results have been averaged over 100 point clouds collected by a 64 channel LiDAR with roughly 200k points each. The DSOR filter is, on average, 141 ms or 28\% faster than the state of the art DROR filter.

% Fig. \ref{fig:filter_times_dsor_vs_sor} compares the time taken to filter a point cloud by the DSOR filter with the SOR filter. On average the DSOR filter performed 2 ms faster than the SOR filter.

% \begin{figure}[htbp]
% \centering
% \includegraphics[width=0.475\textwidth]{figures/filter_times_dsor_vs_dror.png}
% \caption{Our DSOR filter is 140 ms faster than the state of the art DROR filter (averaged over 100 point clouds).}
% \label{fig:filter_times_dsor_vs_dror}
% \end{figure}

% Fig. \ref{fig:filter_times_dsor_vs_dror} compares the time taken to filter a point cloud by the DSOR filter with the DROR filter. On average the DSOR filter performed 140 ms faster than the state of the art DROR filter.

\begin{figure}[htbp]
\centering
\includegraphics[width=0.475\textwidth]{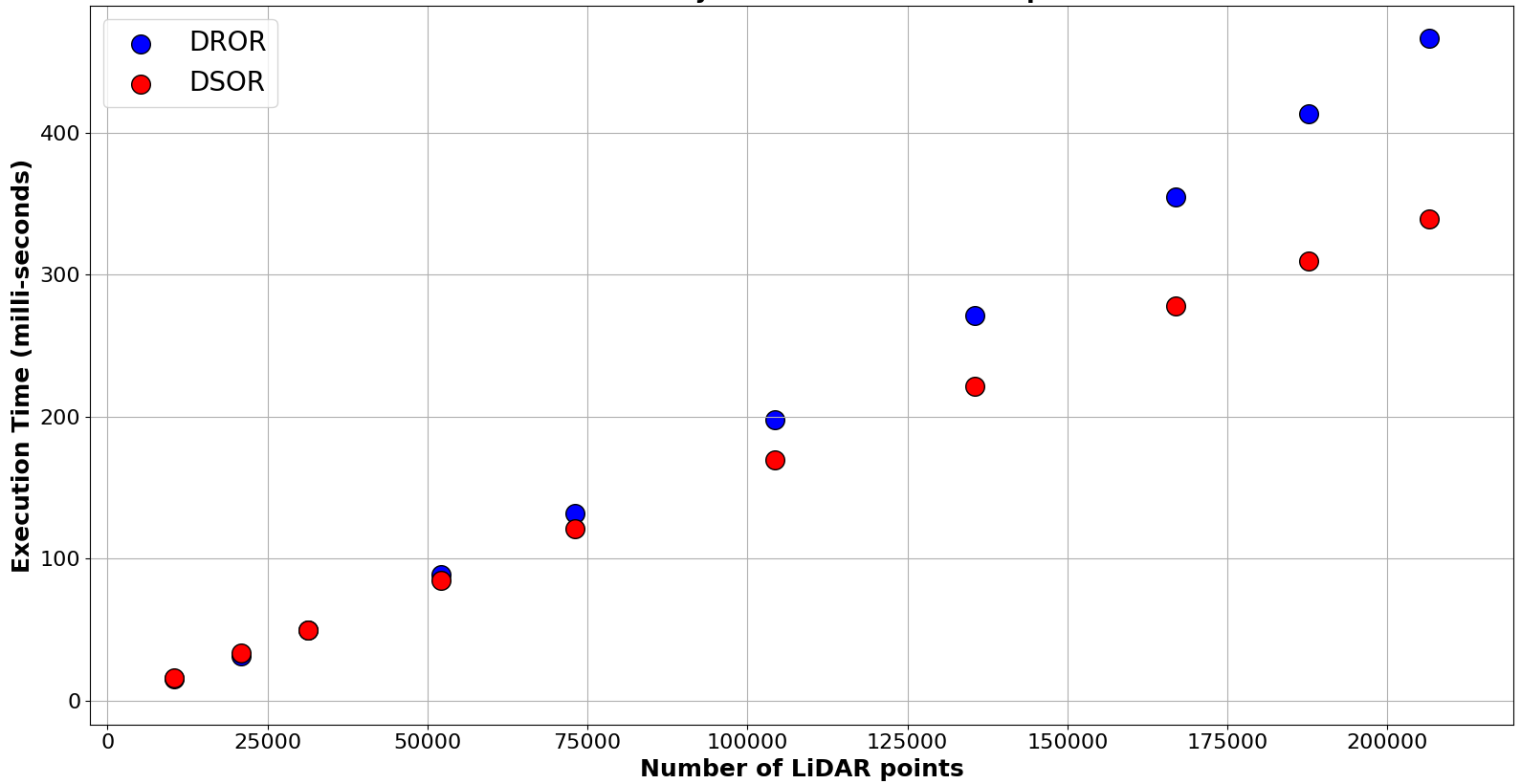}
\caption{Relative time complexity of the DSOR filter comapred to the DROR filter with the size of a point cloud. The DSOR filter scales better than the state of the art DROR filter.}
\label{fig:scalability}
\end{figure}

\subsection{Scalability}
\label{subsec:scalability}
As LiDAR's continue to evolve, sensors with higher point density and number of lasers are expected. State of the art AV's also concatenate returns from multiple LiDAR's for increased visibility and dense spatial sampling. This increases the number of points being captured and processed in the pipeline. For this reason, filters need to scale with the size of point clouds to ensure adaptation on AV systems. To compare the time taken for filtering by the DSOR filter and DROR filter, multiple levels of sub-sampling were applied to a sequence of point clouds from WADS. Fig. \ref{fig:scalability} shows that the DSOR filter has a lower execution time as compared to the DROR filter as the size of a point cloud increases. The DROR filter uses a radius search and we speculate that the DROR filter has an approximate time complexity of $O(N^{2/3})$ \cite{lu2021curve}. A curve fit on Fig. \ref{fig:scalability} is consistent with this assessment. The DSOR filter uses a k-nearest neighbor search and has a lower time complexity of $O(N log(N))$ implying that our filter will scale better with the size of LiDAR point clouds.

\section{conclusion}
\label{sec:conclusion}

Adverse weather conditions negatively affect perception systems used in autonomous vehicles. In particular, LiDAR point clouds suffer from false detections (both positive and negative) introduced by falling rain and snow. Until now, a lack of datasets focused on inclement winter weather has limited the development of AV's to good clear weather conditions. In this work, we introduced the Winter Adverse Driving dataSet (WADS), a dense, point-wise labeled dataset featuring severe winter weather. We also propose two class labels for falling snow and accumulated snow useful for AV tasks like object detection, localization and mapping, and semantic and panoptic segmentation in adverse weather. In the future we would like to increase the size of labeled data and provide annotated images and possibly RADAR data to enable data fusion in winter weather.

We also present the Dynamic Statistical Outlier Removal (DSOR) filter for de-noising snow from LiDAR point clouds. We show that our filter outperforms the state of the art DROR filter by achieving a 4\% higher recall and 28\% lower execution time, without optimization. The DSOR filter has a lower time complexity enabling it to scale better with the size of point clouds. Further improvements in performance may be realized by using fast clustering and voxel-subsampling as shown in \cite{balta2018fast}.

% References
\bibliographystyle{ieeetr}
\bibliography{refs}

\begin{thebibliography}{10}

\bibitem{park2020}
J.-I. Park, J.~Park, and K.-S. Kim, ``Fast and accurate desnowing algorithm for
  lidar point clouds,'' {\em IEEE Access}, vol.~8, pp.~160202--160212, 2020.

\bibitem{bos2020autonomy}
J.~P. Bos, D.~Chopp, A.~Kurup, and N.~Spike, ``Autonomy at the end of the
  earth: an inclement weather autonomous driving data set,'' in {\em Autonomous
  Systems: Sensors, Processing, and Security for Vehicles and Infrastructure
  2020}, vol.~11415, p.~1141507, International Society for Optics and
  Photonics, 2020.

\bibitem{autoware}
S.~Kato, S.~Tokunaga, Y.~Maruyama, S.~Maeda, M.~Hirabayashi, Y.~Kitsukawa,
  A.~Monrroy, T.~Ando, Y.~Fujii, and T.~Azumi, ``Autoware on board: Enabling
  autonomous vehicles with embedded systems,'' in {\em 2018 ACM/IEEE 9th
  International Conference on Cyber-Physical Systems (ICCPS)}, pp.~287--296,
  2018.

\bibitem{behley2019semantickitti}
J.~Behley, M.~Garbade, A.~Milioto, J.~Quenzel, S.~Behnke, C.~Stachniss, and
  J.~Gall, ``Semantickitti: A dataset for semantic scene understanding of lidar
  sequences,'' in {\em Proceedings of the IEEE/CVF International Conference on
  Computer Vision}, pp.~9297--9307, 2019.

\bibitem{caesar2020nuscenes}
H.~Caesar, V.~Bankiti, A.~H. Lang, S.~Vora, V.~E. Liong, Q.~Xu, A.~Krishnan,
  Y.~Pan, G.~Baldan, and O.~Beijbom, ``nuscenes: A multimodal dataset for
  autonomous driving,'' in {\em Proceedings of the IEEE/CVF conference on
  computer vision and pattern recognition}, pp.~11621--11631, 2020.

\bibitem{huang2020}
X.~Huang, P.~Wang, X.~Cheng, D.~Zhou, Q.~Geng, and R.~Yang, ``The apolloscape
  open dataset for autonomous driving and its application,'' {\em IEEE
  Transactions on Pattern Analysis and Machine Intelligence}, vol.~42,
  p.~2702–2719, Oct 2020.

\bibitem{bijelic2020}
M.~Bijelic, T.~Gruber, F.~Mannan, F.~Kraus, W.~Ritter, K.~Dietmayer, and
  F.~Heide, ``Seeing through fog without seeing fog: Deep multimodal sensor
  fusion in unseen adverse weather,'' in {\em The IEEE/CVF Conference on
  Computer Vision and Pattern Recognition (CVPR)}, June 2020.

\bibitem{pitropov2020}
M.~Pitropov, D.~E. Garcia, J.~Rebello, M.~Smart, C.~Wang, K.~Czarnecki, and
  S.~Waslander, ``Canadian adverse driving conditions dataset,'' {\em The
  International Journal of Robotics Research}, vol.~40, p.~681–690, Dec 2020.

\bibitem{pfeuffer2018optimal}
A.~Pfeuffer and K.~Dietmayer, ``Optimal sensor data fusion architecture for
  object detection in adverse weather conditions,'' 2018.

\bibitem{michaud2015}
S.~Michaud, J.-F. Lalonde, and P.~Giguere, ``Towards characterizing the
  behavior of lidars in snowy conditions,'' in {\em 2015 IEEE/RSJ International
  Conference on Intelligent Robots and Systems (IROS)}, vol.~7, 2015.

\bibitem{charron2018denoising}
N.~Charron, S.~Phillips, and S.~L. Waslander, ``De-noising of lidar point
  clouds corrupted by snowfall,'' in {\em 2018 15th Conference on Computer and
  Robot Vision (CRV)}, pp.~254--261, IEEE, 2018.

\bibitem{xu2021spg}
Q.~Xu, Y.~Zhou, W.~Wang, C.~R. Qi, and D.~Anguelov, ``Spg: Unsupervised domain
  adaptation for 3d object detection via semantic point generation,'' 2021.

\bibitem{rasmussen1999estimation}
R.~M. Rasmussen, J.~Vivekanandan, J.~Cole, B.~Myers, and C.~Masters, ``The
  estimation of snowfall rate using visibility,'' {\em Journal of Applied
  Meteorology and Climatology}, vol.~38, no.~10, pp.~1542--1563, 1999.

\bibitem{roy20}
G.~Roy, X.~Cao, R.~Bernier, and G.~Tremblay, ``Physical model of snow
  precipitation interaction with a 3d lidar scanner,'' {\em Appl. Opt.},
  vol.~59, pp.~7660--7669, Sep 2020.

\bibitem{xie2020}
Y.~Xie, J.~Tian, and X.~X. Zhu, ``Linking points with labels in 3d: A review of
  point cloud semantic segmentation,'' {\em IEEE Geoscience and Remote Sensing
  Magazine}, vol.~8, no.~4, pp.~38--59, 2020.

\bibitem{sakaridis2018}
C.~Sakaridis, D.~Dai, and L.~Van~Gool, ``Semantic foggy scene understanding
  with synthetic data,'' {\em International Journal of Computer Vision},
  vol.~126, p.~973–992, Mar 2018.

\bibitem{heinzler_cnn-based_2020}
R.~Heinzler, F.~Piewak, P.~Schindler, and W.~Stork, ``{CNN}-{Based} {Lidar}
  {Point} {Cloud} {De}-{Noising} in {Adverse} {Weather},'' {\em IEEE Robotics
  and Automation Letters}, vol.~5, pp.~2514--2521, Apr. 2020.
\newblock Conference Name: IEEE Robotics and Automation Letters.

\bibitem{geiger2013vision}
A.~Geiger, P.~Lenz, C.~Stiller, and R.~Urtasun, ``Vision meets robotics: The
  kitti dataset,'' {\em The International Journal of Robotics Research},
  vol.~32, no.~11, pp.~1231--1237, 2013.

\bibitem{duan2020lowcomplexity}
Y.~Duan, C.~Yang, H.~Chen, W.~Yan, and H.~Li, ``Low-complexity point cloud
  filtering for lidar by pca-based dimension reduction,'' 2020.

\bibitem{rusu2011pcl}
R.~B. Rusu and S.~Cousins, ``3d is here: Point cloud library (pcl),'' in {\em
  2011 IEEE International Conference on Robotics and Automation}, pp.~1--4,
  2011.

\bibitem{Geiger2013IJRR}
A.~Geiger, P.~Lenz, C.~Stiller, and R.~Urtasun, ``Vision meets robotics: The
  kitti dataset,'' {\em International Journal of Robotics Research (IJRR)},
  2013.

\bibitem{lu2021curve}
Y.~Lu, L.~Cheng, T.~Isenberg, C.-W. Fu, G.~Chen, H.~Liu, O.~Deussen, and
  Y.~Wang, ``Curve complexity heuristic kd-trees for neighborhood-based
  exploration of 3d curves,'' in {\em Computer Graphics Forum}, vol.~40,
  pp.~461--474, Wiley Online Library, 2021.

\bibitem{balta2018fast}
H.~Balta, J.~Velagic, W.~Bosschaerts, G.~De~Cubber, and B.~Siciliano, ``Fast
  statistical outlier removal based method for large 3d point clouds of outdoor
  environments,'' {\em IFAC-PapersOnLine}, vol.~51, no.~22, pp.~348--353, 2018.

\end{thebibliography}

\end{document}